\documentclass{article}
\usepackage{apacite}

\usepackage[preprint]{nips_2018}

\usepackage[utf8]{inputenc} 
\usepackage[T1]{fontenc}    
\usepackage{hyperref}       
\usepackage{url}            
\usepackage{booktabs}       
\usepackage{amsfonts}       
\usepackage{nicefrac}       
\usepackage{microtype}      
\usepackage{amsmath} 
\usepackage{amssymb}
\usepackage{graphicx}
\usepackage{subfigure}
\usepackage{geometry}

\title{Training Generative Adversarial Networks Via Turing Test}

\author{
  Jianlin Su \\
  School of Mathematics\\
  Sun Yat-sen University\\
  Guangdong, China \\
  \texttt{bojone@spaces.ac.cn} \\
}

\begin{document}

\maketitle

\begin{abstract}
  In this article, we introduce a new mode for training Generative Adversarial Networks (GANs). Rather than minimizing the distance of evidence distribution $\tilde{p}(x)$ and the generative distribution $q(x)$, we minimize the distance of $\tilde{p}(x_r)q(x_f)$ and $\tilde{p}(x_f)q(x_r)$. This adversarial pattern can be interpreted as a Turing test in GANs. It allows us to use information of real samples during training generator and accelerates the whole training procedure. We even find that just proportionally increasing the size of discriminator and generator, it succeeds on 256x256 resolution without adjusting hyperparameters carefully.
\end{abstract}

\section{Reviews of GANs}

GANs has been developed a lot since Goodfellow's fist work \citep{Goodfellow2014Generative}. The main idea of GANs is to train a generator $G(z)$ such that the generative distribution
\begin{equation}q(x) = \int \delta(x - G(z))q(z)dz\end{equation}
will be a good approximation of the evidence distribution $\tilde{p}(x)$, while $q(z)$ is a prior distribution which will be standard normal distribution usually. Generally, the current GANs aim to minimize the distribution distance of $\tilde{p}(x)$ and $q(x)$.

\subsection{Standard GANs}

Here a series of GANs which are based on the Goodfellow's fist work are called Standard GANs (SGANs). Firstly, we fix generator and train a discriminator $T(x)$ by the following goal
\begin{equation}\mathop{\arg\max}_{T}\mathbb{E}_{x\sim \tilde{p}(x)}[\log \sigma(T(x))] + \mathbb{E}_{x\sim q(x)}[\log(1-\sigma(T(x)))]\label{eq:sgan-d}\end{equation}
whose $\sigma(x) = 1/(1+e^{-x})$ means sigmoid activation. Then we fix discriminator and train a generator $G(z)$ by minimizing
\begin{equation}\mathop{\arg\min}_{G}\mathbb{E}_{x\sim q(x)}[h(T(x))]=\mathop{\arg\min}_{G}\mathbb{E}_{z\sim q(z)}[h(T(G(z)))]\label{eq:sgan-g}\end{equation}
whose $h$ can be any scalar function to make $h(log(t))$ be a convex function of variable $t$. Run two steps alternately and we may get a good generator finally.

Using variational method, we can show that the optimum solution of $\eqref{eq:sgan-d}$ is
\begin{equation}\frac{\tilde{p}(x)}{q(x)}=\frac{\sigma(T(x))}{1 - \sigma(T(x))} = e^{T(x)}\end{equation}
replace $T(x)$ in $\eqref{eq:sgan-g}$ with this result, we get
\begin{equation}\begin{aligned}&\mathop{\arg\min}_{G}\mathbb{E}_{x\sim q(x)}\left[h\left(\log\frac{\tilde{p}(x)}{q(x)}\right)\right]\\
=&\mathop{\arg\min}_{G}\int q(x)\left[h\left(\log\frac{\tilde{p}(x)}{q(x)}\right)\right]dx \end{aligned}\end{equation}
Let $f(t)=h(\log(t))$, we can see the essential goal of SGANs is to minimize the $f$-divergence \citep{Nowozin2016f} between $\tilde{p}(x)$ and $q(x)$. Function $f$ is constrained in convex function. Therefore, any function $h$ making $h(\log(t))$ be a convex function is allowed to use, such as $h(t)=-t, h(t)=-\log \sigma(t), h(t)=\log (1-\sigma(t))$, which lead to the following loss of generator:
\begin{equation}-T(x),\quad - \log \sigma(T(x)),\quad \log(1 - \sigma(T(x)))\end{equation}

\subsection{Wasserstein GANs}

An important breakthrough in GANs is Wasserstein GANs (WGANs, \cite{Arjovsky2017Wasserstein}). Compared with SGANs, WGANs can improve the stability of learning and get rid of problems like mode collapse. The main idea of WGANs is to minimize the Wasserstein distance of $\tilde{p}(x)$ and $q(x)$, rather than $f$-divergence in SGANs. The Wasserstein distance
\begin{equation}W(\tilde{p}(x), q(x)) = \inf_{\gamma\in \Pi(\tilde{p}(x), q(x))} \mathbb{E}_{(x,y)\sim \gamma}\Vert x-y\Vert \end{equation}
is an excellent metric of two distribution. $\gamma\in \Pi(\tilde{p}(x), q(x))$ means $\gamma$ is any joint distribution of variable $x$ and $y$ whose marginal distributions are $\tilde{p}(x)$ and $q(y)$. With a dual transformation, Wasserstein distance can be rewritten as
\begin{equation}W(\tilde{p}(x), q(x)) = \sup_{\Vert T\Vert_L\leq 1} \mathbb{E}_{x\sim \tilde{p}(x)}[T(x)] -  \mathbb{E}_{x\sim q(x)}[T(x)]\label{eq:wgan-d}\end{equation}
whose $T(x)$ is a scalar function and $\Vert T\Vert_L$ is Lipschitz norm of function $T$:
\begin{equation}\Vert T\Vert_L = \max_{x\neq y} \frac{|T(x)-T(y)|}{\Vert x - y\Vert}\end{equation}

With these foundations, we can train the generator as a min-max game under the Wasserstein distance:
\begin{equation}\mathop{\arg\min}_{G}\mathop{\arg\max}_{T,\Vert T\Vert_L\leq 1} \mathbb{E}_{x\sim \tilde{p}(x)}[T(x)] -  \mathbb{E}_{x\sim q(x)}[T(x)]\end{equation}
The first $\arg\max$ attempts to acquire a approximate function of Wasserstein distance and the second $\arg\min$ attempts to minimize the Wasserstein distance of $\tilde{p}(x)$ and $q(x)$.

One difficulty of WGANs is how to impose Lipschitz constraint $\Vert T\Vert_L\leq 1$ on $T$, which currently has serveral solutions: weight clipping \citep{Arjovsky2017Wasserstein}, gradient penalty \citep{Gulrajani2017Improved} and spectral normalization \citep{Miyato2018Spectral}.

\subsection{Problems}

GANs has achieved a great success but there are still some problems waiting to be solved. 

The distinct one is that training of GANs will be very unstable on large-scale datasets, such as 256x256 images and higher. Simply increasing the size of discriminator and generator can always not achieve this goal. It always needs certain tricks and well-designed hyperparameters for discriminator and generator, and even needs a large amount of computing resources \citep{karras2017progressive,brock2018large,Xue2018Variational}.

\section{A New GANs' Mode}

There are two things in common between SGANs and WGANs: 1. They both attempts to minimize one kind of distribution distance between $\tilde{p}(x)$ and $q(x)$; 2. While updating generator, only fake samples from generative distribution is available. 

So the updating of generator depends on whether discriminator can remember characteristics of real samples or not. In other words, generator just improve its production by the memory of discriminator, using no signal of real samples directly. It may be too hard to discriminator and lower the convergence rate of generator.

Here we demonstrate a new mode of GANs: to minimize distance of $\tilde{p}(x_r)q(x_f)$ and $\tilde{p}(x_f)q(x_r)$. This idea can make real images available while updating generator and can be integrated into all the current GANs. It is a new thought to train all generative models rather than one specific GANs.

\subsection{Under SGANs}

Define two joint distributions
\begin{equation}P(x_r, x_f) = \tilde{p}(x_r)q(x_f),\quad Q(x_r, x_f) = \tilde{p}(x_f)q(x_r)\end{equation}
now we want to minimize the distance of $P(x_r, x_f)$ and $Q(x_r, x_f)$. Regard $(x_r, x_f)$ as one whole random variable, and from $\eqref{eq:sgan-d}$ we get
\begin{equation}\begin{aligned}&\mathop{\arg\max}_{T}\mathbb{E}_{(x_r, x_f)\sim P(x_r, x_f)}[\log \sigma(T(x_r, x_f))] + \mathbb{E}_{(x_r, x_f)\sim Q(x_r, x_f)}[\log(1-\sigma(T(x_r, x_f)))]\\
=&\mathop{\arg\max}_{T}\mathbb{E}_{(x_r, x_f)\sim \tilde{p}(x_r)q(x_f)}[\log \sigma(T(x_r, x_f))+\log(1-\sigma(T(x_f, x_r)))]\end{aligned}\label{eq:tsgan-d}\end{equation}
Then from $\eqref{eq:sgan-g}$ we have
\begin{equation}\begin{aligned}&\mathop{\arg\min}_{G}\mathbb{E}_{(x_r, x_f)\sim Q(x_r, x_f)}[h(T(x_r, x_f))]\\
=&\mathop{\arg\min}_{G}\mathbb{E}_{x_r\sim \tilde{p}(x_r), x_f\sim q(x_f)}[h(T(x_f, x_r))]\\
=&\mathop{\arg\min}_{G}\mathbb{E}_{x_r\sim \tilde{p}(x_r), z\sim q(z)}[h(T(G(z), x_r))]
\end{aligned}\label{eq:tsgan-g}\end{equation}
Therefore, we can train a generative model by alternately running the following two steps:
\begin{equation}\begin{aligned}&\mathop{\arg\max}_{T}\mathbb{E}_{(x_r, x_f)\sim \tilde{p}(x_r)q(x_f)}[\log \sigma(T(x_r, x_f))+\log(1-\sigma(T(x_f, x_r)))]\\
&\qquad\qquad\qquad\mathop{\arg\min}_{G}\mathbb{E}_{x_r\sim \tilde{p}(x_r), x_f\sim q(x_f)}[h(T(x_f, x_r))]\end{aligned}\label{eq:tsgan-dg}\end{equation}
A natural choice of $h$ leads to
\begin{equation}\begin{aligned}&\mathop{\arg\max}_{T}\mathbb{E}_{(x_r, x_f)\sim \tilde{p}(x_r)q(x_f)}[\log \sigma(T(x_r, x_f))+\log(1-\sigma(T(x_f, x_r)))]\\
&\mathop{\arg\max}_{G}\mathbb{E}_{(x_r, x_f)\sim \tilde{p}(x_r)q(x_f)}[\log (1-\sigma(T(x_r, x_f)))+\log \sigma(T(x_f, x_r))]\end{aligned}\label{eq:tsgan-dg-2}\end{equation}

\subsection{Under WGANs}

Corresponding to $\eqref{eq:wgan-d}$, we can estimate Wasserstein distance between $P(x_r, x_f)$ and $Q(x_r, x_f)$ by
\begin{equation}\begin{aligned}&W(P(x_r, x_f),Q(x_r, x_f)) \\
=& \sup_{\Vert T\Vert_L\leq 1} \mathbb{E}_{(x_r, x_f)\sim P(x_r, x_f)}[T(x_r, x_f)] -  \mathbb{E}_{(x_r, x_f)\sim Q(x_r, x_f)}[T(x_r, x_f)]\\
=& \sup_{\Vert T\Vert_L\leq 1} \mathbb{E}_{(x_r, x_f)\sim \tilde{p}(x_r)q(x_f)}[T(x_r, x_f)-T(x_f, x_r)]\end{aligned}\label{eq:twgan-d}\end{equation}
Hence we can train a generative model by a new min-max game:
\begin{equation}\mathop{\arg\min}_{G}\mathop{\arg\max}_{T,\Vert T\Vert_L\leq 1} \mathbb{E}_{(x_r, x_f)\sim \tilde{p}(x_r)q(x_f)}[T(x_r, x_f)-T(x_f, x_r)]\label{eq:twgan-dg}\end{equation}
It is a really pretty result, which allows us to use an exactly symmetrical target to train discriminator and generator.

\subsection{Relate to Turing Test}

There is a very intuitive interpretation for minimizing the distance of $\tilde{p}(x_r)q(x_f)$ and $\tilde{p}(x_f)q(x_r)$: Turing test \citep{Turing1995Computing}.

As we known, Turing test is a test of a machine's ability to exhibit intelligent behavior equivalent to, or indistinguishable from, that of a human. The tester communicates with both the robot and the human in unpredictable situations. If the tester fails to distinguish the human from the robot, we can say the robot has (in some aspects) human intelligence.

How about it in GANs? If we sample $x_r$ from real distribution $\tilde{p}(x_r)$ and $x_f$ from fake distribution $q(x_f)$, then mix them. Can we identify where they come from? That is, how much difference between $\tilde{p}(x_r)q(x_f)$ and $\tilde{p}(x_f)q(x_r)$? A good generator means we have $\tilde{p}(x)\approx q(x)$ everywhere, so we can not distinguish $\tilde{p}(x_r)q(x_f)$ and $\tilde{p}(x_f)q(x_r)$, so does $(x_r, x_f)$ and $(x_f, x_r)$.

Therefore, to minimize the distance of $\tilde{p}(x_r)q(x_f)$ and $\tilde{p}(x_f)q(x_r)$ is like a Turing test in GANs. We mix real samples and fake samples such that discriminator has to distinguish them by pairwise comparison and generator has to improve itself by pairwise comparison.

We call GANs in this mode as Turing GANs (T-GANs), correspondingly, $\eqref{eq:tsgan-dg}$ as T-SGANs and $\eqref{eq:twgan-dg}$ as T-WGANs.

\section{Related Works}

Both $\eqref{eq:tsgan-dg}$ and $\eqref{eq:twgan-dg}$ allow optimizer to obtain the signal of real samples directly to update generator. Formally, compared with SGANs and WGANs, the discriminator of T-GANs is two-variables function which needs both real and fake sample as inputs. It means that discriminator needs a pairwise comparison to make a reasonable judgement. 

This idea fistly occurs in RSGANs \citep{Jolicoeurmartineau2018The}. Our result can be regarded as an expansion of RSGANs. Just define $T(x_r, x_f) \triangleq T(x_r) - T(x_f)$ in $\eqref{eq:tsgan-dg-2}$, with $1 - \sigma(x) = \sigma(-x)$ we can obtain RSGANs:
\begin{equation}\begin{aligned}&\mathop{\arg\max}_{T}\mathbb{E}_{(x_r, x_f)\sim \tilde{p}(x_r)q(x_f)}[\log \sigma(T(x_r) - T(x_f))]\\
&\mathop{\arg\max}_{G}\mathbb{E}_{(x_r, x_f)\sim \tilde{p}(x_r)q(x_f)}[\log \sigma(T(x_f) - T(x_r))]\end{aligned}\label{eq:rsgan-dg}\end{equation}
RSGANs have demonstrate some potential to improve GANs and we will demonstrate more efficient and sustainable progress of T-GANs at the section $\ref{sec:experiments}$.

However, RSGANs is not the first GANs which make real samples available during training generator. As far as I know, the first one is Cramer GANs \citep{Bellemare2017The}, which is based on energy distance:
\begin{equation}\mathop{\arg\min}_{G}\mathop{\arg\max}_{E,\Vert E\Vert_L\leq 1} \mathbb{E}_{x_{r,1},x_{r,2}\sim \tilde{p}(x_r),\,x_{f,1},x_{f,2}\sim q(x_f)}[f(E(x_{r,1}),E(x_{r,2}),E(x_{f,1}),E(x_{f,2}))]\label{eq:cramer-dg}\end{equation}
whose $E$ is an encoder network and
\begin{equation}f(x_1, x_2, y_1, y_2)= \Vert x_1 - y_2\Vert + \Vert y_1 - x_2\Vert - \Vert x_1 - x_2\Vert - \Vert y_1 - y_2\Vert\end{equation}
and
\begin{equation}\iiiint p(x_1)p(x_2)q(y_1)q(y_2)f(x_1, x_2, y_1, y_2) dx_1 dx_2 dy_1 dy_2\end{equation}
is called energy distance of $p(x)$ and $q(x)$. Cramer GAN is not a perfect and complete inference framework of generative models. In fact it seems like a empirical model and it does not work well on large-scale datasets. It need more samples for echo updating iteration which is computation intensive.

\section{Experiments\label{sec:experiments}}

Our experiments are conducted on CelebA HQ dataset \citep{liu2015faceattributes} and cifar10 dataset \citep{krizhevsky2009learning}. We test both $\eqref{eq:tsgan-dg}$ and $\eqref{eq:twgan-dg}$ on CelebA HQ of 64x64, 128x128 and 256x256 resolution. cifar10 is an additional auxiliary experiment to demonstrate T-GANs work better than current GANs.

Code was written in Keras \citep{chollet2015keras} and available in my repository\footnote{\url{https://github.com/bojone/T-GANs}}. The architectures of models were modified from DCGANs \citep{Radford2015Unsupervised}. And models were trained using Adam optimizer \citep{Kingma2014Adam} with learning rate 0.0002 and momentum 0.5. 

Experiments on 64x64 and 128x128 resolution were run on a GTX 1060 and experiments on 256x256 resolution were run on a GTX 1080Ti.

\subsection{Design of Discriminator}

In theory, any neural network with double inputs $x_r, x_f$ can be used as $T(x_r, x_f)$. But for simplicity, inspired by RSGANs, we design $T(x_r, x_f)$  into the following form:
\begin{equation}T(x_r, x_f) \triangleq D(E(x_r) - E(x_f))\end{equation}
whose $E(\cdot)$ is an encoder for input image and $D(\cdot)$ is a multilayer perception with hidden difference vector of $E(x_r), E(x_f)$ as input and a scalar as output. It can also be regarded as a relativistic discriminator comparing the hidden features of $x_r$ and $x_f$, rather than comparing the final scalar output in RSGANs.

If we use T-SGANs $\eqref{eq:tsgan-dg}$, no constraints for $T$ theoretically. But as we known, gradient vanishing usually occurs in SGANs and spectral normalization is an effective strategy to prevent it. Therefore, spectral normalization has been a popular trick to be added into discriminator, no matter SGANs or WGANs, so do T-SGANs and T-WGANs.

Our experiment demonstrates $\eqref{eq:tsgan-dg}$ and $\eqref{eq:twgan-dg}$ have similar performance while spectral normalization is applied on their discriminator $T(x_r, x_f)$.

\subsection{Result Analysis}

On 64x64 resolution's experiments, we find T-SGANs and T-WGANs has a faster convergence rate than popular GANs, such as DCGANs, DCGANs-SN, WGANs-GP, WGANs-SN, RSGANs. 

On 128x128 resolution, we find most popular GANs does not work or only work under very particular hyperparameters and convergences unsteadily, but T-GANs still work well and have the same convergence rate as on 64x64 resolution. 

We even find that just proportionally increasing the size of discriminator and generator, it succeeds on 256x256 resolution. It is very incredible that training a generative adversarial model on high resolution does not need to adjust hyperparameters carefully under T-GANs framework.

\subsubsection{Faster Convergence Rate}

\begin{figure}
  \centering
  \includegraphics[width=17cm]{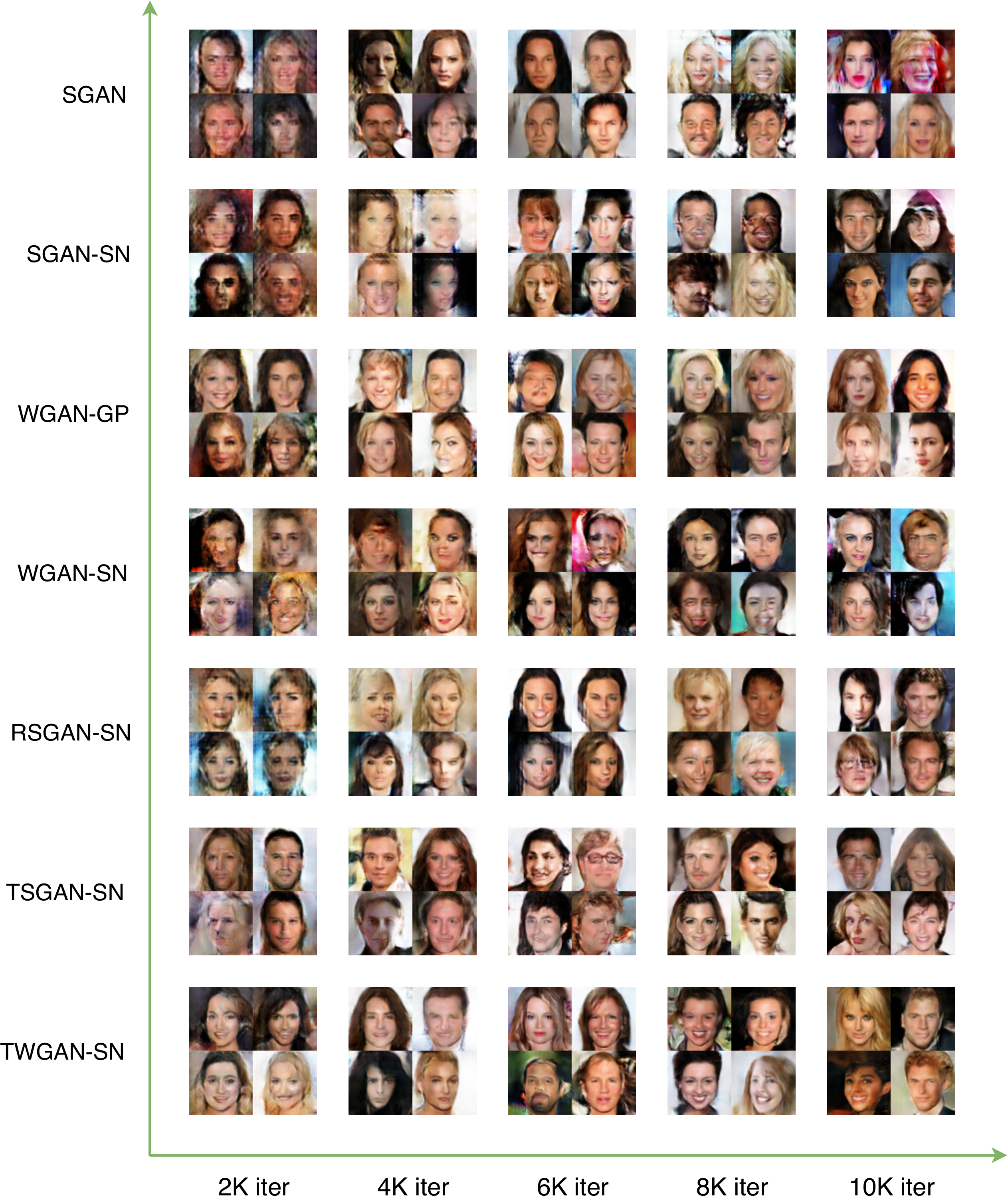}
  \caption{Comparison of convergence rate of different GANs on 64x64 CelebA. T-GANs converges almost twice as fast as other GANs. "-SN" means spectral normalization is added into discriminator.}
  \label{fig:rate}
\end{figure}

Figure $\ref{fig:rate}$ shows comparison of convergence rate of different GANs. All of these GANs has same architecture of discriminator and generator. And they actually have same final performance but different convergence rate. We found that T-SGANs and T-WGANs converges almost twice as fast as other GANs. other GANs need about 20k iterations to achieve the same performance as T-GANs do in 10k iterations.

It needs to be pointed out that WGAN-GP seems to have a similar performance like T-GANs but actually it needs more time for echo iteration. During echo iteration, we update discriminator 5 times and update generator 1 times while training WGAN-GP and WGAN-SN, but update discriminator 1 times and update generator 2 times while training other GANs (including T-SGANs and T-WGANs).

\begin{figure}
  \centering
  \includegraphics[width=17cm]{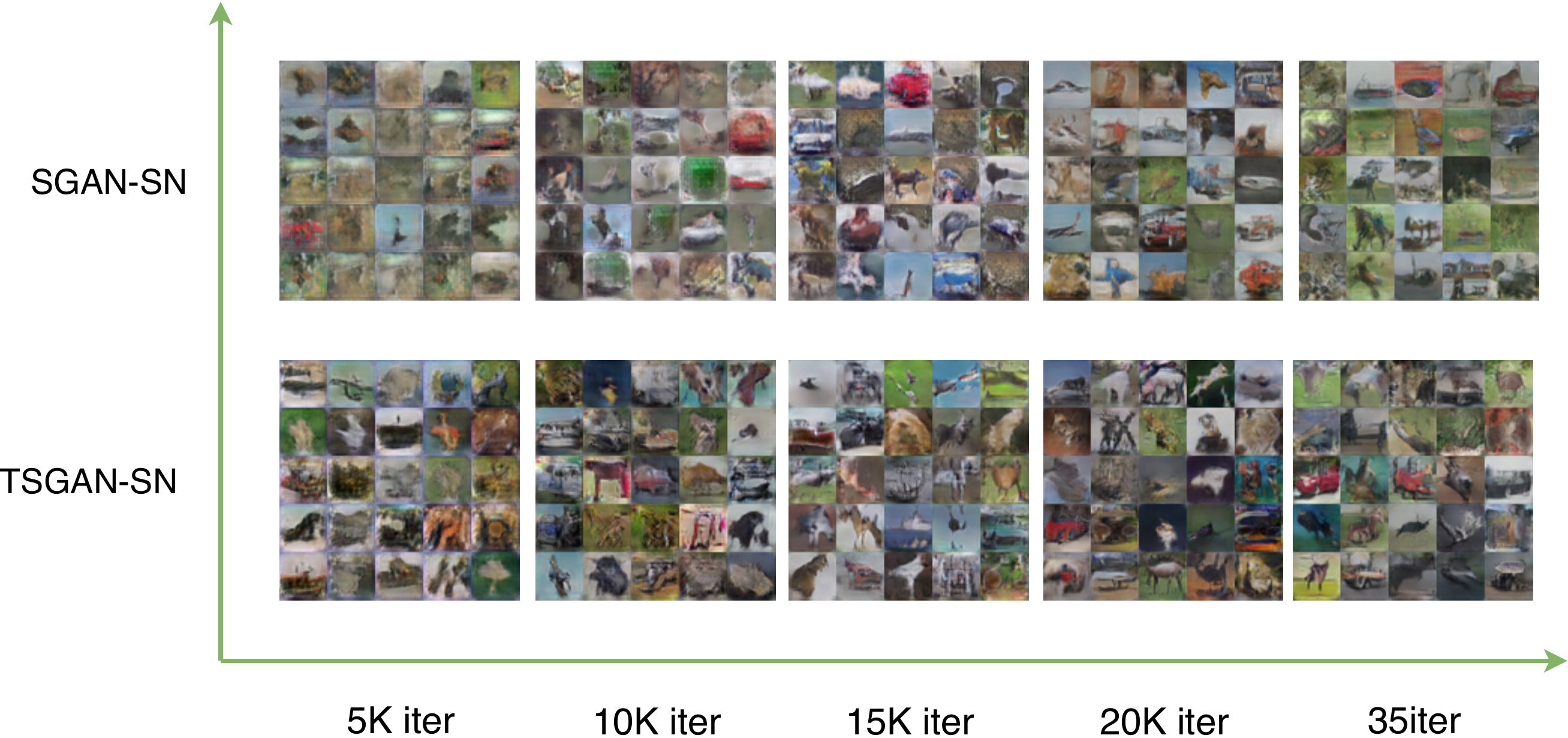}
  \caption{Comparison of convergence rate of different GANs on cifar10. It suggests that GANs under mode of Turing test has a better convergence than conventional. WGAN-SN performs like SGAN-SN and TWGAN-SN performs like TSGAN-SN, so we just show the result of SGAN-SN and TSGAN-SN.}
  \label{fig:cifar10}
\end{figure}

Experiments on cifar10 also demonstrates this conclusion further (Figure $\ref{fig:cifar10}$).

\subsubsection{High Quality Generation}

Now we will focus attention on high quality generation. On 128x128 resolution, we compare serveral GANs but few of them work well. Frechet Inception Distance(FID, \cite{Heusel2017GANs}) is used as a quantitative index to evaluate these models. Table $\ref{tab:128}$ demonstrates T-GANs still work well while increasing resolution, only need to expand the size of models.

\begin{table}
  \renewcommand\arraystretch{1.5}
  \caption{Final results of serveral GANs on 128x128 resolution.}
  \label{tab:128}
  \centering
  \begin{tabular}{c|ccccc}
    \hline
    \hline
    & SGAN-SN      & WGAN-SN & RSGAN-SN  & TSGAN-SN & TWGAN-SN \\
    \hline
   IMG & \parbox[c]{62pt}{\vspace{0.1cm}\includegraphics[width=64pt]{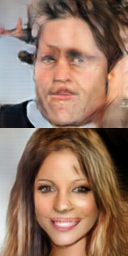}\vspace{0.1cm}} 
       & \parbox[c]{62pt}{\includegraphics[width=64pt]{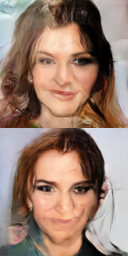}} 
       & \parbox[c]{62pt}{\includegraphics[width=64pt]{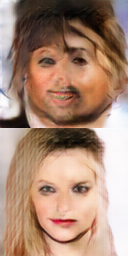}} 
       & \parbox[c]{62pt}{\includegraphics[width=64pt]{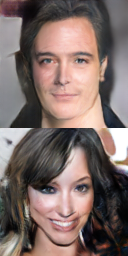}}
       & \parbox[c]{62pt}{\includegraphics[width=64pt]{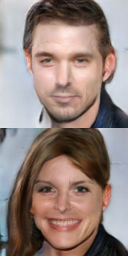}} \\
    \hline
   FID & 65.78 & 66.63 & 98.87 & 46.92 & 48.88\\
    \hline
    \hline
  \end{tabular}
\end{table}

\begin{figure}
  \centering
  \includegraphics[width=16cm]{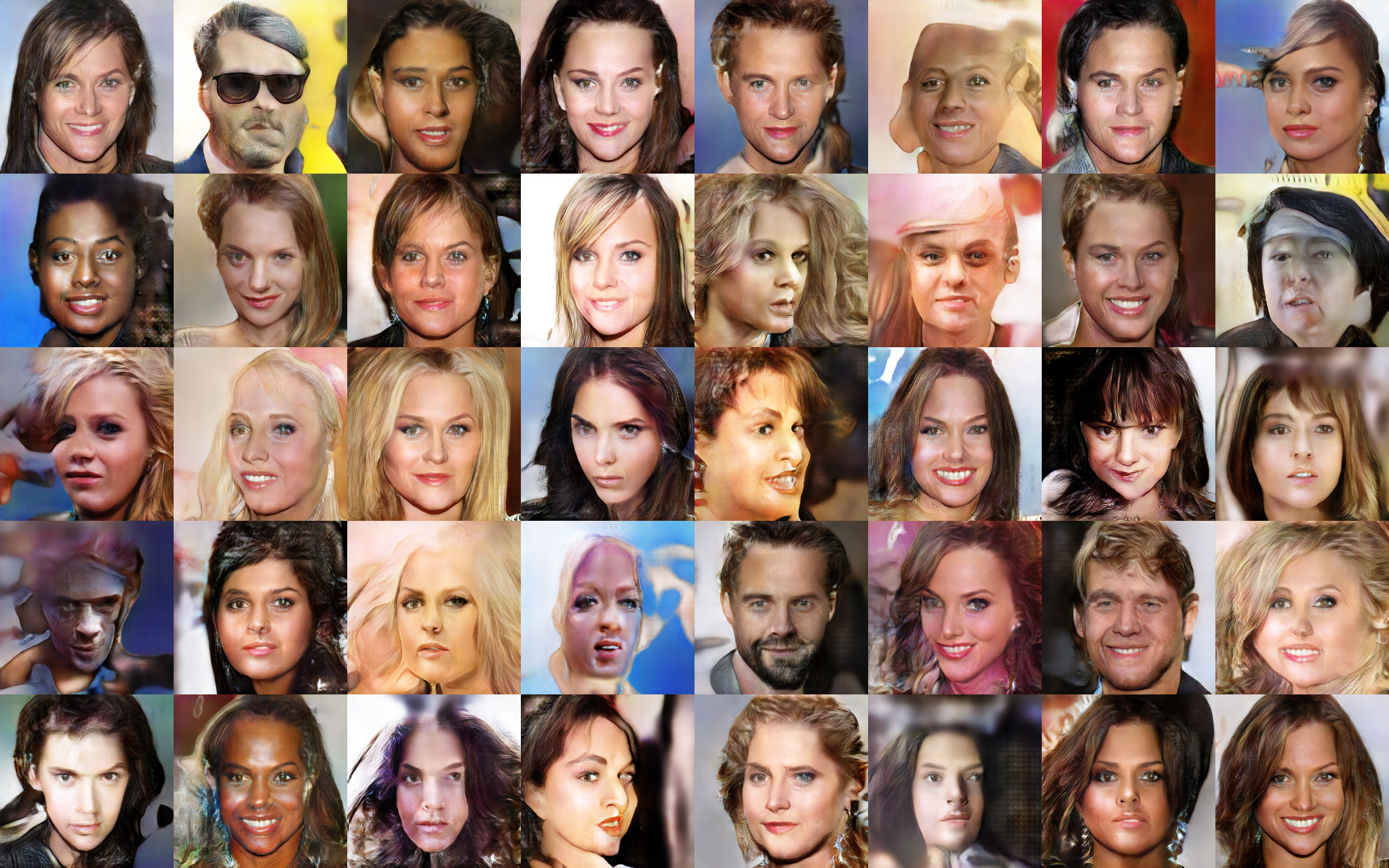}
  \caption{Random samples of T-SGANs on 256x256 resolution.}
  \label{fig:256}
\end{figure}

We also test T-GANs on 256x256 resolution and T-GANs also work well but all of others fail to do that (Figure $\ref{fig:256}$). It is worth mentioning that no matter 64x64, 128x128 or 256x256 resolution, T-GANs would achieve a good performance after 12000 iterations. That is to say, large-scale does not affect the convergence of the T-GANs.

\section{Conclusion}

In this paper, we propose a new adversarial mode for training generative models called T-GANs. This adversarial pattern can be interpreted as a Turing test in GANs. It is a guiding ideology for training GANs rather than a specific GANs model. It can be integrated with current popular GANs such SGANs and WGANs, leading to T-SGANs and T-WGANs.

Our experiments demonstrate that T-GANs have good and stable performance on dataset varying from small scale to large scale. It suggests the signal of real samples is really important during updating generator in GANs. However, the mechanism of T-GANs to improve stability and convergence rate remains to be explored further.

\medskip

\small

\bibliographystyle{apacite}
\bibliography{tgan.bib}

\newgeometry{left=1.8cm,top=2cm,right=1.8cm}

\begin{figure}
  \centering
  \includegraphics[width=16cm]{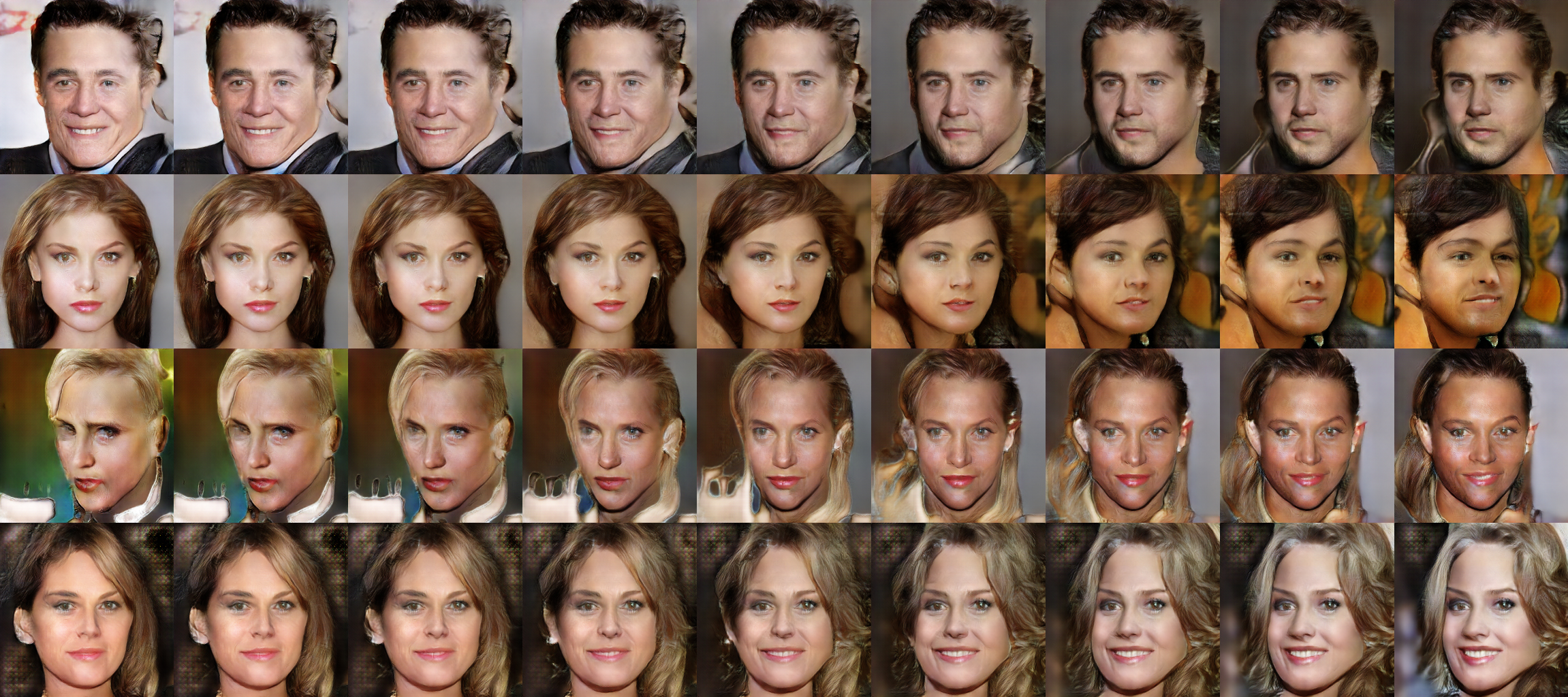}
  \caption{Random interpolation of T-SGANs on 256x256 resolution.}
  \label{fig:256-inter}
\end{figure}

\begin{figure}
  \centering
  \setlength{\abovecaptionskip}{0.1cm}
  \subfigure[Random samples from SGAN-SN]{
    \includegraphics[height=8.5cm]{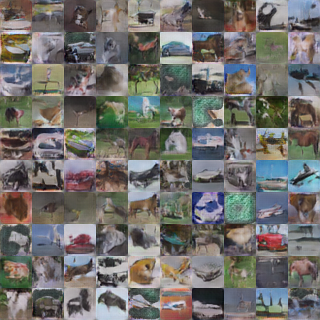}}\hspace{0.25cm}
  \subfigure[Random samples from TSGAN-SN]{
    \includegraphics[height=8.5cm]{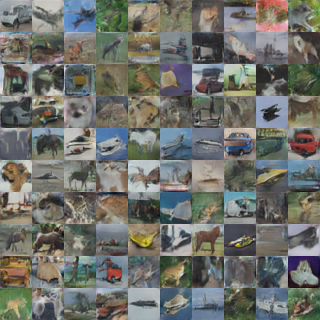}}
  \caption{Random samples from cifar10 (60K iteratons).}
  \label{fig:Comparasion}
\end{figure}

\restoregeometry

\end{document}